\documentclass[runningheads]{llncs}
\usepackage{graphicx}
\usepackage{caption}
\usepackage{subcaption}
\usepackage{multirow}
\usepackage{amsmath}
\usepackage{nicefrac}
\usepackage{hyperref}
\usepackage{xcolor} 
\usepackage{colortbl}
\usepackage{array} 

\usepackage{tikz}
\usepackage{pgfplots}
\usepackage{float}
\pgfplotsset{compat=1.18}
\pgfplotsset{paraviewStyle/.style={%
hide axis,scale only 
axis,height=0pt,width=0pt,xmin=0,xmax=1,ymin=0,ymax=1,domain=0:1, 
point meta=f(x),
}}

\pgfplotsset{
    colormap={w_r}{
        color(0cm)=(white);
        color(2cm)=(black)
    }
}

\begin{document}
\title{From Structured to Unstructured: \\ A Comparative Analysis of Computer Vision and Graph Models in solving Mesh-based PDEs \thanks{This research has been funded by the Federal Ministry for Economic Affairs and Climate Action (BMWK) within the project "KI-basierte Topologieoptimierung elektrischer Maschinen (KITE)" (19I21034C).}}

\titlerunning{Solving Mesh-based PDEs}
\author{Jens Decke\inst{1,2}\orcidID{0000-0002-7893-1564} \and
Olaf Wünsch\inst{2}\orcidID{0000-0002-7295-8862}  \and \\
Bernhard Sick\inst{1}\orcidID{0000-0001-9467-656X} \and 
Christian Gruhl\inst{1}\orcidID{0000-0001-9838-3676}
}
\authorrunning{J. Decke et al.}
%
\institute{Intelligent Embedded Systems, University of Kassel, 34121 Kassel, Germany\\ \url{https://www.uni-kassel.de/eecs/ies/} 
\and Fluid Dynamics, University of Kassel, 34125 Kassel, Germany \\ \url{https://www.uni-kassel.de/go/fluiddynamics} 
\\
\email{\{jdecke, wuensch, bsick, cgruhl\}@uni-kassel.de}
}
\maketitle
%
\begin{abstract}
This article investigates the application of computer vision and graph-based models in solving mesh-based partial differential equations within high-performance computing environments. Focusing on structured, graded structured, and unstructured meshes, the study compares the performance and computational efficiency of three computer vision-based models against three graph-based models across three data\-sets. The research aims to identify the most suitable models for different mesh topographies, particularly highlighting the exploration of graded meshes, a less studied area. Results demonstrate that computer vision-based models, notably U-Net, outperform the graph models in prediction performance and efficiency in two (structured and graded) out of three mesh topographies. The study also reveals the unexpected effectiveness of computer vision-based models in handling unstructured meshes, suggesting a potential shift in methodological approaches for data-driven partial differential equation learning. The article underscores deep learning as a viable and potentially sustainable way to enhance traditional high-performance computing methods, advocating for informed model selection based on the topography of the mesh.

\end{abstract}
\section{Introduction} \label{introduction}%
    Partial differential equations (PDEs) are central to modeling a wide range of complex physical phenomena in various scientific and engineering disciplines, including fluid dynamics and electrodynamics. Traditionally, solving these equations in detailed simulations necessitates high-performance computing (HPC) resources~\cite{wu2022sustainable,katal2023energy,Yunus2024A}. However, the immense computational power of HPC systems leads to substantial energy consumption, raising environmental concerns due to the associated greenhouse gas emissions. In an era where sustainable computing is becoming increasingly crucial, our research explores a new data-driven approach to lower the computational demands of these simulations.

\begin{figure}[b!]
        \centering
        \begin{subfigure}[b]{0.32\textwidth}
            \includegraphics[width=\textwidth]{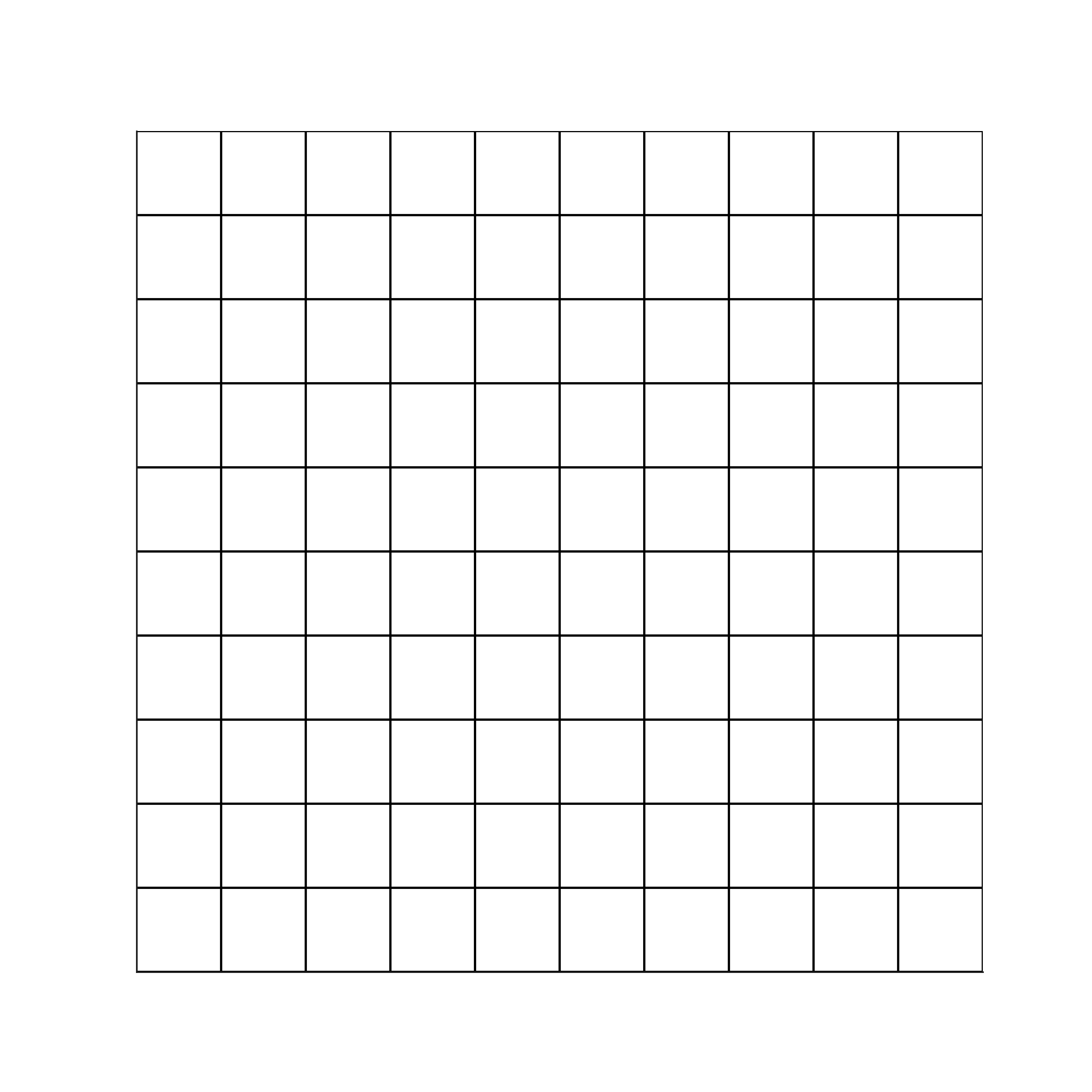}
            \caption{Structured mesh}
            \label{fig:rugular}
        \end{subfigure}
        \hfill
        \begin{subfigure}[b]{0.32\textwidth}
            \includegraphics[width=\textwidth]{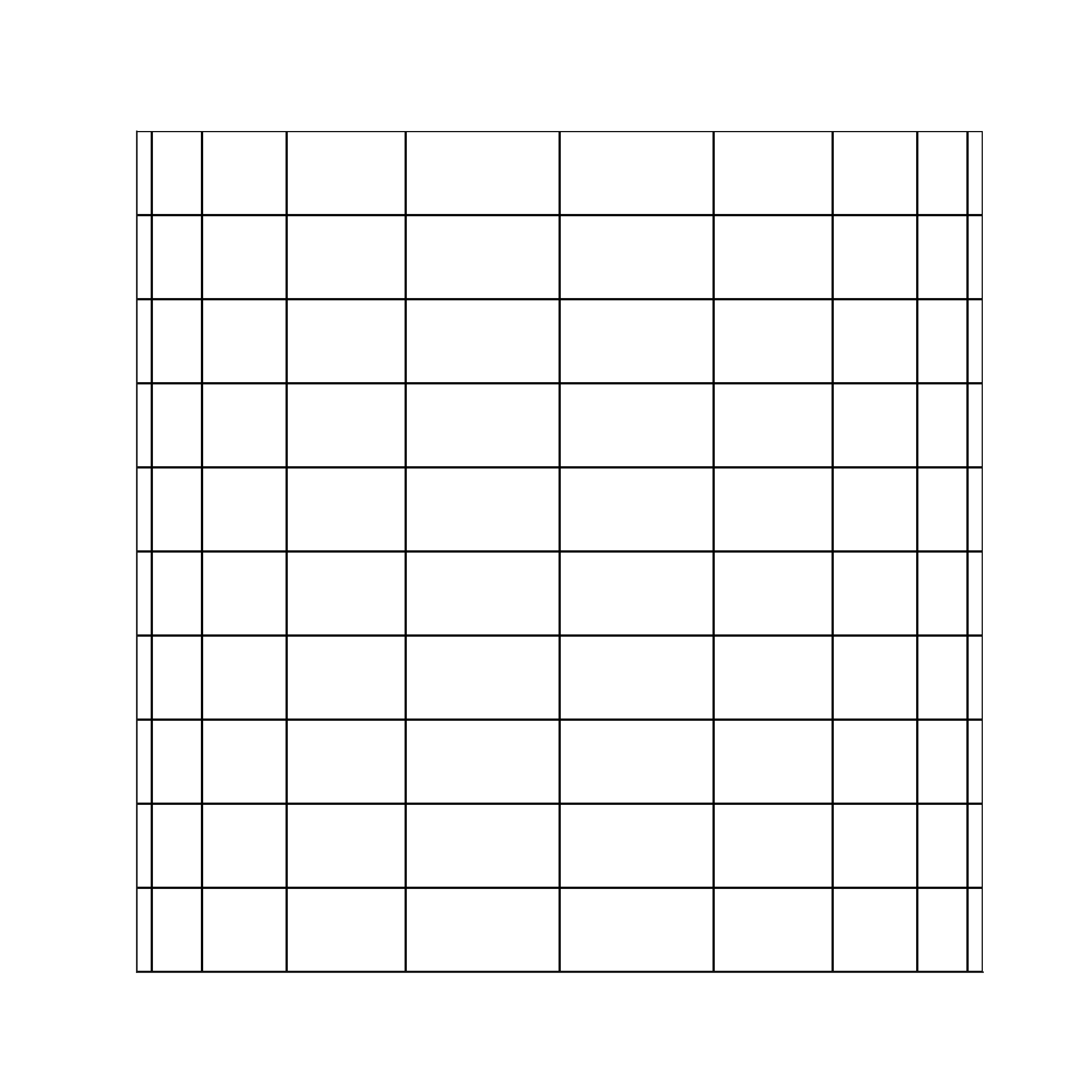}
            \caption{Graded mesh}
            \label{fig:graded}
        \end{subfigure}
        \hfill
        \begin{subfigure}[b]{0.32\textwidth}
            \includegraphics[width=\textwidth]{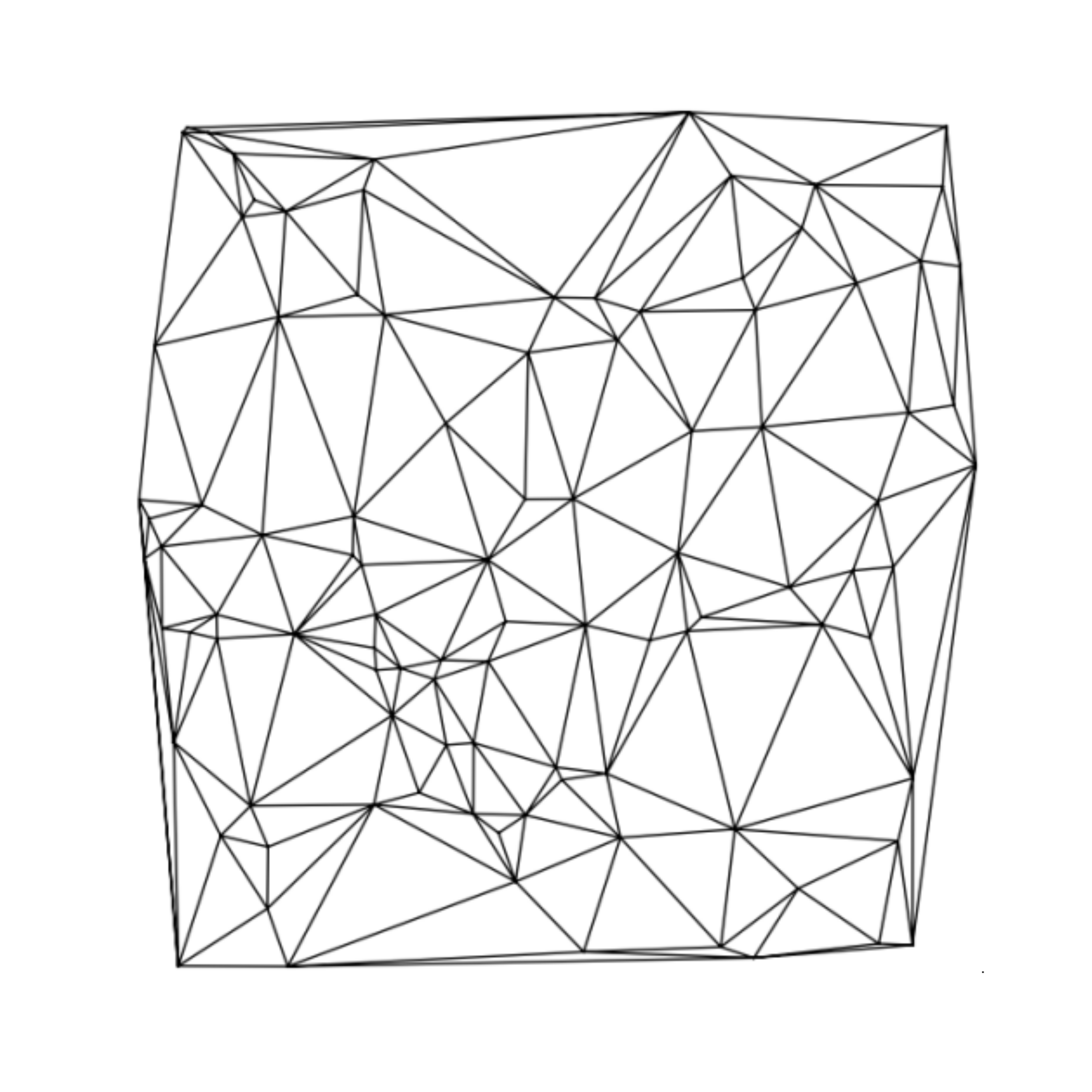}
            \caption{Unstructured mesh}
            \label{fig:irregular}
        \end{subfigure}
        \caption{Three different mesh topographies. \textbf{(a)} displays a structured mesh, which is believed to be advantageous for CV-based models due to its conformity with pixel images. \textbf{(b)} illustrates a graded mesh challenges the boundary between structured and unstructured meshes. In \textbf{(c)} depicts an unstructured mesh exemplifies the domain of graph-based models, optimized for irregular connectivity. This study evaluates whether CV-based or graph-based models are better suited for graded mesh topography.}
        \label{fig:mesh_structures}
    \end{figure}

    This article investigates the application of advanced deep learning models from the fields of computer vision (CV) and graph learning in solving mesh-based PDEs in an HPC context. Figure~\ref{fig:mesh_structures} provides illustrative examples of this study's three distinct mesh topographies. We focus on six models of these two fields, tailored to three specific datasets: \emph{Darcy Flow}~\cite{takamoto2022PDEBench} on a structured mesh (cf. Figure~\ref{fig:rugular}), a \emph{U-bend}~\cite{decke2023dataset} with a graded structured mesh (cf. Figure~\ref{fig:graded}) hereafter referred to as graded mesh, and an \emph{electric motor}~\cite{Botache2023Enhancing} with an unstructured mesh (cf. Figure~\ref{fig:irregular}). We investigate the ability of CV-based architectures and graph-based models to accurately predict solutions and determine which model can offer the most energy-efficient alternatives to traditional HPC methods. CV-based models are known for their capabilities in analyzing structured data, while graph-based ones are effective in unstructured data scenarios. Although graded meshes have the inherent advantage of maintaining neighborhood relations of a structured mesh while also allowing for higher resolution in areas of interest, they have found limited attention in existing research. We fill this gap by determining whether CV-based or graph-based models are more suitable for these mesh topographies. 

    We detail the adaptation of CV-based and Graph-based models for these diverse datasets and evaluate their computational efficiency and performance. By showcasing the potential of these models to reduce both computational demands and environmental impact, our study advances sustainable HPC solutions for solving PDE with data-driven methods. 

    One popular and widely adopted way to reduce the number of evaluations needed for finding good-enough samples employs deep active design optimization (DADO)~\cite{decke2023dado}. This advanced organic computing approach, effectively minimizes evaluation counts through strategic, self-aware acquisition methods. This method seeks to optimize computational efficiency by strategically choosing the most informative points for evaluation, utilizing a numerical simulation as an expert. This reduces the computational resources required by minimizing the total number of numerical simulations needed. Our investigation focuses on identifying a model that offers quick, efficient, and high-quality results suitable for incorporation into the DADO framework. This model serves as a surrogate for evaluating queries, offering evidently faster inference times than traditional numerical simulations. 
    
    Beyond design optimization, these surrogate models will find critical use in self-aware microsystems (SAM)~\cite{gruhl2022self}, where the rapid inference capabilities of the surrogates are essential. By employing these surrogates within SAM, we embed knowledge about physical phenomena into the real-time decision-making processes of robots, a feature unachievable with time-consuming numerical simulations. This integration enables SAM to respond adaptively to environmental changes swiftly, bypassing the extensive computation times closely resembling the core idea to make decision at runtime~\cite{muller2011organic}.
    The code~\footnote{\url{https://gitlab.uni-kassel.de/uk023894/cv_vs_graph}} used in our study is provided to facilitate further research based on our findings, along with configuration files to reproduce our experiments. We summarize our key contributions (\textbf{C1}-\textbf{C4}) as follows:
    
    \begin{itemize}
        \item[] \textbf{C1:} We incorporate state-of-the-art transformer based foundation models.
        \item[] \textbf{C2:} We apply the latest graph-based models designed for mesh-based PDEs. 
        \item[] \textbf{C3:} We deliver an in-depth comparison between CV and graph-based models
        \item[] \textbf{C4:} We offer insights to model selection for surrogates in DADO and SAM.
    \end{itemize}

    The structure of this article is as follows: Section~\ref{related_work} reviews existing literature, focusing on the application of CV-based and graph-based models in solving physical simulations of PDEs. Section~\ref{methodology} details the six models and three datasets employed in our research and provides a comprehensive outline of our evaluation methodology. Section~\ref{results} presents our findings and discusses the implications and significance of our results. The article concludes with Section~\ref{conclusion}, which reflects our findings and proposes directions for future research in this evolving field.
    
\section{Related Work}\label{related_work}%
    CV-based models such as Convolutional Neural Networks, with their roots in image processing, excel in identifying spatial patterns and relationships, making them inherently suited for structured mesh environments~\cite{defferrard2016Convolutional,raonic2023convolutional}. This tendency arises because information in pixel images is evenly distributed, a characteristic also found in structured mesh configurations. Li et al.~\cite{li2023transformer} introduced the idea of using transformers, which is built on self-attention, cross-attention, and multilayer perceptrons to solve PDEs. They also applied their model on structured (uniform) and unstructured (non-uniform) mesh datasets. In contrast to Li et al. we incorporate the Self-Distillation with No Labels~\cite{oquab2024dinov} (DINOv2) model which leverages self-supervised learning to generate robust visual representations without relying on labeled data. By employing self-distillation, DINOv2 uncovers intricate patterns in images, enabling broad applicability in vision tasks without the extensive need for annotated datasets. This highlights DINOv2s as a promising approach to reduce reliance on large labeled datasets. In addition, the Data-Efficient Image Transformer~\cite{touvron2021training} (DeiT) utilizes the transformer architecture to achieve impressive results with limited image data. DeiT's novel training strategy, featuring distillation tokens, allows it to efficiently learn from smaller datasets, positioning it as a substantial development in data-efficient computing. DeiT demonstrates the potential of transformers to operate effectively in scenarios where data is scarce. 
    
    Graph-based models have been increasingly applied to solve PDEs in mesh-based simulations, demonstrating particular effectiveness in handling the complexities of unstructured meshes, which are pivotal for representing intricate geometric shapes~\cite{bronstein2017geometric,brandstetter2022message,gao2022graph}. Their flexibility in adapting to the variable nature of these meshes contrasts with the preference for structured meshes in numerical simulations (which are valued for their computational efficiency and precision), enabling fast and more accurate problem resolution. In this article, we intend to utilize three graph-based models. The first model is the Bi-Stride Multi-Scale Graph Neural Network (BSMS)~\cite{Cao2023Efficient}, designed to handle varying scales and complexities in mesh topography. The second model is the Mesh Graph Net (MGN)~\cite{pfaff2021learning}, it focuses on capturing intricate relationships within mesh-based data. Lastly, the Mesh Graph Net + Tree (MGTN)~\cite{ripken2023multiscale}, proposed by Ripken et al., is developed to process hierarchical tree edges efficiently. 

    Recent studies provide insights into the performance of neural operators across different mesh topographies. Ripken et al.~\cite{ripken2023multiscale} compared graph-based models against a U-Net and a Fourier neural operator on the \emph{Darcy Flow} dataset from Takamoto et al.~\cite{takamoto2022PDEBench}, revealing that graph-based models did not surpass the U-Net in scenarios involving a equidistant mesh resolution of 128x128 cells. Additionally, their exploration extended to the \emph{Electric Motor} dataset with an unstructured mesh provided by Botache et al.~\cite{Botache2023Enhancing}, where CV-based models were not considered, emphasizing the reliance on graph-based models for such complex geometries. Their experiments demonstrate that the graph rewiring strategy in their model not only boosts the performance of baseline methods but also achieves state-of-the-art results in certain tasks, indicating a considerable advancement in the efficiency and performance of neural PDE operators.
     
    A mesh topography that incorporate strong grading near boundaries enhances local resolution in these areas, as illustrated in Figure~\ref{fig:graded}. This approach commonly observed in fluid dynamics, introduces additional challenges. These graded meshes, designed to capture steep gradients near walls, are exemplified by a dataset from Decke et al.~\cite{decke2023dataset}, showcasing a \emph{U-bend Flow} with variable mesh cell distribution to enhance resolution in critical areas. Such datasets underline the need for neural network models to adapt to non-equidistant mesh resolution, a challenge not addressed in current literature.

\section{Methodology and Data}\label{methodology}%
    This research aims to evaluate advanced deep learning models from the field of CV and graph learning on structured, graded and unstructured meshes. By focusing on their adaptability and performance in learning numerical simulations, particularly in the less-explored area of graded meshes, the study seeks to provide insights for choosing suitable models. This could improve PDE solution strategies in computationally demanding areas and advance methodologies in this field. In this section, we first describe the models applied, then introduce the three datasets and outline our experimental setup.
\subsection{Models}
The six models used in our experiments are divided into CV-based and graph-based models.

\textbf{CV-based} focuses on leveraging architectural innovations for enhanced visual task performance. These models, known for their ability to process extensive image datasets through advanced feature extraction techniques, are implemented using resources from Hugging Face~\cite{wolf2020transformers} with their default parameters.
    \begin{itemize}
        \item The \textbf{U-Net}~\cite{ronneberger2015UNet} revolutionized medical image segmentation with its encoder-decoder architecture, employing skip connections and attention mechanisms for precise feature extraction and localization. Its design is pivotal for integrating low-level details with high-level features, ensuring detailed and high-performance image tasks.
    
        \item \textbf{Self-Distillation with No Labels} (DINOv2)~\cite{oquab2024dinov} adopts a self-supervised learning approach, distilling knowledge within the model to learn rich visual representations without labeled data. This method allows DINOv2 to excel in tasks requiring a deep understanding of complex image content, making it a versatile tool for various image-based analyses.
    
        \item \textbf{Data-Efficient Image Transformer} (DeiT)~\cite{touvron2021training} showcases the transformer model's capability to handle visual data efficiently, particularly in limited data scenarios. DeiT's data-efficient training strategy enables competitive performance against larger models, emphasizing its adaptability to data constraints.
    \end{itemize}
For the DINOv2 and the DeiT, we utilize these models as encoders due to their robust capabilities in capturing comprehensive visual representations. For both models, we employ their corresponding base size configuration. Their encoded outputs are further processed by a straightforward decoder network that includes four blocks of upsampling layers, convolutional layers, and LeakyReLU activation functions. This decoder design is aimed at efficiently reconstructing the targets from the encoded representations.

\textbf{Graph-based} models, adept at navigating complex geometrical and physical data through mesh-to-graph conversions and multi-scale analysis, leverage implementations from Ripken et al.~\cite{ripken2023multiscale} with default parameters.
    \begin{itemize}
        \item \textbf{Bi-Stride Multi-Scale Graph Neural Network} (BSMS)~\cite{Cao2023Efficient} focuses on multi-scale representation in mesh-based simulations, integrating a bi-stride mechanism to analyze physical phenomena at multiple scales accurately. Its approach to selectively aggregating information enhances simulation detail and efficiency, particularly in structural analysis and fluid dynamics.
    
        \item \textbf{Mesh Graph Net} (MGN)~\cite{pfaff2021learning} merges mesh-based geometries with GNNs for efficient learning and processing of complex geometrical data. It excels in simulating physical systems across various domains by mapping mesh topography into graph representations, capturing both local and global interactions effectively.
    
        \item \textbf{Mesh Graph Net + Tree} (MGTN)~\cite{ripken2023multiscale}, an extension of MGN, targets the specific challenges of PDEs on large meshes with a novel graph rewiring technique. This method improves global interactions on irregular meshes by effectively aggregating information across scales, showcasing MGTN's advanced capability in handling complex simulations.
    \end{itemize}
    The hyperparameters are chosen according to the experiments of Ripken~et al.~\cite{ripken2023multiscale} considering the 8 closest neighbors of the graph.

\subsection{Datasets}
\begin{figure}[b!]
        \centering
        \begin{subfigure}[b]{0.32\textwidth}
            \includegraphics[width=.8\textwidth]{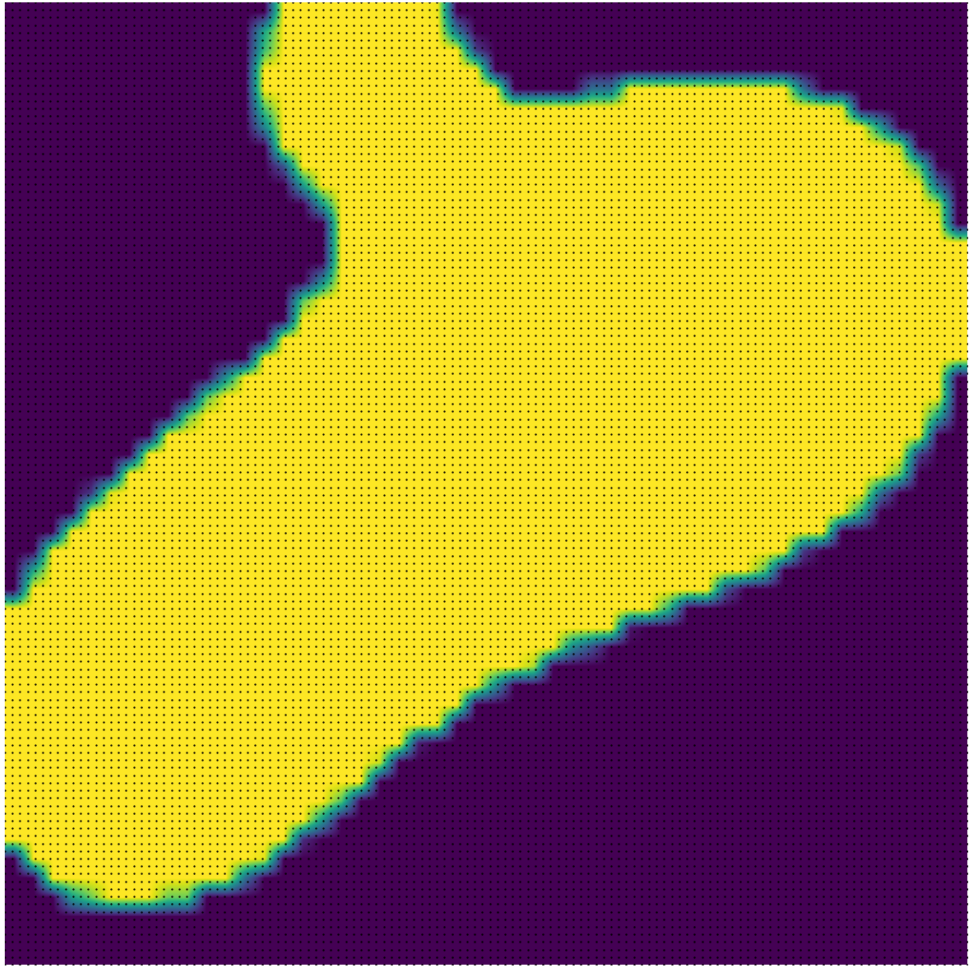}
            \caption{Darcy Flow}
            \label{fig:rugular_darcy}
        \end{subfigure}
        \hfill
        \begin{subfigure}[b]{0.32\textwidth}
            \includegraphics[width=\textwidth]{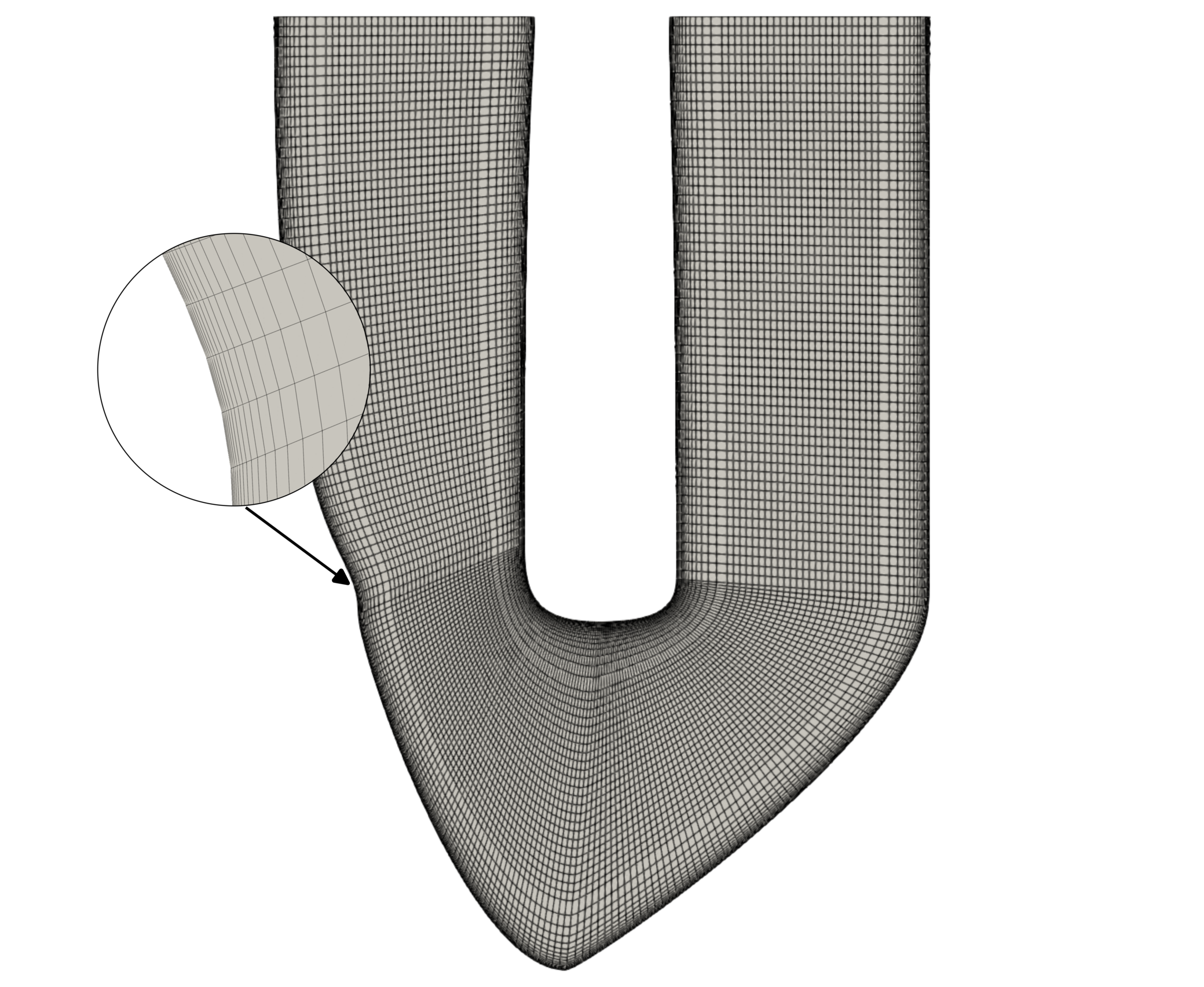}
            \caption{U-bend Flow}
            \label{fig:graded_ubend}
        \end{subfigure}
        \hfill
        \begin{subfigure}[b]{0.32\textwidth}
            \includegraphics[width=\textwidth]{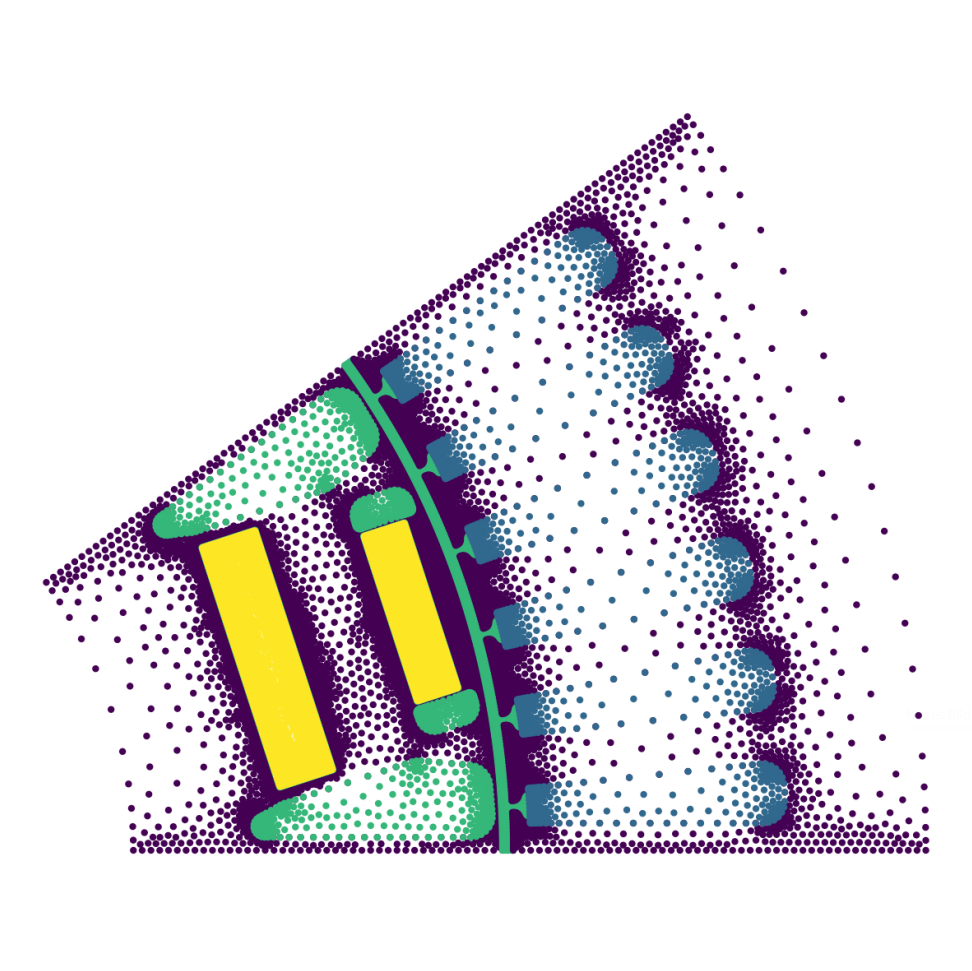}
            \caption{Electric Motor}
            \label{fig:irregular_motor}
        \end{subfigure}
        \caption{Input samples from the three datasets used in this article. \textbf{(a)} corresponds to a structured mesh, \textbf{(b)} to a graded mesh with a finer resolution towards the walls, and \textbf{(c)} to an unstructured mesh. See also Figure~\ref{fig:mesh_structures} for reference.}
        \label{fig:samples}
    \end{figure}

This subsection provides insights into the various datasets and their origins, offering a closer look at the different types of data used. Figure~\ref{fig:samples} illustrates a representative sample from each dataset, demonstrating the input provided to the models. Specifically, Figure~\ref{fig:rugular_darcy} presents an input sample from the \emph{Darcy Flow} dataset (i.e. the hydraulic conductivity). In contrast, Figure~\ref{fig:graded_ubend} showcases a design of a U-bend, and Figure~\ref{fig:irregular_motor} reveals the material distribution within a sample from the \emph{Electric Motor} dataset.

\textbf{a.) \emph{Darcy Flow} dataset:} This dataset is characterized by a structured mesh, designed to simulate fluid flow through porous media~\cite{takamoto2022PDEBench}. The structured mesh ensures uniformity in data distribution. The mesh has a resolution of 128x128 meshcells. The Darcy dataset is derived from the porous media equation, representing a generalized PDE describing fluid flow through porous materials. This PDE is a second-order, nonlinear parabolic PDE due to diffusion terms and possible nonlinearity in permeability or fluid properties. Equation~\eqref{eq:darcy} shows the porous media equation, based on Darcy's law, which links fluid \emph{velocity} to pressure gradients $\nabla p(\mathbf{x}, t)$, assuming an incompressible fluid and steady-state conditions. The equation incorporates spatial variations $\mathbf{x}$ in hydraulic conductivity $K$ and includes a constant source term to account for external forces $\mathbf{f}$. A dynamic method is used to approach steady-state solutions through temporal evolution, showcasing the dataset's foundation on both theoretical and practical aspects of fluid dynamics in porous media. This methodology ensures the Darcy dataset accurately reflects the complexities involved in simulating real-world porous media flow scenarios.

    \begin{equation}
       \frac{\partial p(\mathbf{x}, t)}{\partial t} - \nabla \cdot (K(\mathbf{x}) \nabla p(\mathbf{x}, t)) = \mathbf{f}, 
       \label{eq:darcy}
    \end{equation}

Understanding and modeling fluid behavior in porous media is crucial for various sectors, including groundwater movement, agriculture, filtration, construction, biomedicine, geothermal energy, environmental cleanup, and carbon capture, spanning diverse scientific and industrial fields. This dataset is provided by~\cite{takamoto2022PDEBench} it contains 10,000 samples in total. The hydraulic conductivity is meant to be used as the input (cf. Figure~\ref{fig:rugular_darcy}) and the \emph{pressure} $p$ is the solution variable, which we use as the model output. As Ripken et al. did before, we used the dataset with a constant source term $\mathbf{f}=\beta=1$. In contrast to them we chose a smaller dataset size of 5,000 to expedite training and provide a more realistic comparison, given the typical scarcity of data samples in PDE solving using deep learning strategies. 

\textbf{b.) \emph{U-bend Flow} dataset:} Employing a graded mesh, the \emph{U-bend Flow} dataset simulates fluid flow in curved geometries, with the help of a graded mesh~\cite{decke2023dataset}. This gradation in mesh topography poses a unique challenge for model performance and efficiency, testing the models' adaptability to the complexity of the graduation of the mesh. Figure~\ref{fig:graded_ubend} illustrates an exemplary U-bend design, featuring its graded computational mesh. Accompanying the design, a detailed view of the mesh adjacent to the wall is also provided, where the pronounced grading is clearly visible. The PDE which describes the fluid flow through a U-bend is the Navier-Stokes equation it is a second-order, nonlinear PDE due to viscosity terms and nonlinearity from convective acceleration. In Equation~\eqref{eq:ubend}-\eqref{eq:ubend2} the adiabatic, time-independent (steady-state) Navier-Stokes equations with linear relations between stress and strain rates (Newtonian fluid; $\eta = konst.$) with constant density $\rho$ (incompressible) is depicted:
    \begin{align}
        \rho \left( \mathbf{u} \cdot \nabla \mathbf{u} \right) &= -\nabla p + \eta \nabla^2 \mathbf{u} + \mathbf{f}, \label{eq:ubend} \\
        \nabla \cdot \mathbf{u} &= 0 \label{eq:ubend2}
    \end{align}
The term $\rho (\mathbf{u} \cdot \nabla \mathbf{u})$ represents the convective acceleration, which is the change in \emph{velocity} $\mathbf{u}$ as fluid moves along a streamline, $ -\nabla p$ represents the force due to pressure gradients. Viscous forces are denoted by $\eta \nabla^2 \mathbf{u}$, Where $\eta$ is the dynamic viscosity, characteristic of Newtonian fluids where the stress is directly proportional to the strain rate. External forces such as gravity acting on the fluid are accounted by $\mathbf{f}$. Equation \eqref{eq:ubend2} ensures the incompressibility of the fluid, stating that the fluid density remains constant within the flow field.

Fluid flow through U-bend shapes is common across diverse sectors including chemical plants, power generation facilities, the food industry, as well as automotive and aircraft industry for example. These shapes are integral to piping systems and heat exchangers, optimizing fluid flow and heat transfer. 
Similar to our approach with the Darcy dataset, we limit our analysis to 3000 samples from the graded mesh data. We use the x and y-coordinates of the mesh as the model's input and we try to predict the solution variables \emph{pressure} $p$ and the magnitude of the \emph{velocity} vector $\lVert \mathbf{u} \rVert$. The mesh of the U-bend is characterized by 60 cells across the cross-section and 240 cells along the U-bend, and is effectively treated as a 240x60 array for computational purposes. 
This treatment simplifies the U-bend into a straight channel, a process that inherently ignores the non-equidistant spatial distribution of mesh information. Until now, the application CV-based models to such a scenario has not been explored. Our research aims to bridge this gap by evaluating whether CV-based or graph-based models offer superior performance in scenarios characterized by a strongly graded mesh distribution. Before processing, we pad this array into the dimension of 256x256.

\textbf{c.) \emph{Electric Motor} dataset:} This dataset is provided by Botache et al.~\cite{Botache2023Enhancing}, with its unstructured mesh, represents the most complex geometry, simulating electromagnetic fields in electric motor designs. The irregular topography of the mesh in this dataset tests the model's capabilities in handling non-uniform distributed information in the mesh. Figure~\ref{fig:irregular_motor} displays a design of an electric motor, with colors indicating the various materials used. Iron is represented by the dark purple nodes, air cavities by green, electrical coils by blue, and the motor's magnets by yellow. The Maxwell equations is a first-order, linear PDE in their classical form, describing electromagnetic fields using vectors and operators in mathematical terms. Its time-independent (steady-state) form can be seen in Equations~\eqref{eq:motor}-\eqref{eq:motor4} and is described in the following:

\begin{align}
\nabla \cdot \mathbf{E} &= \nicefrac{\rho}{\varepsilon_0}, \label{eq:motor} \\
\nabla \cdot \mathbf{B} &= 0, \\
\nabla \times \mathbf{E} &= 0, \\
\nabla \times \mathbf{B} &= \mu_0 \mathbf{J} \label{eq:motor4}
\end{align}

 The electric field vector, $\mathbf{E}$, quantifies the force per unit charge, while the magnetic field vector, $\mathbf{B}$, indicates the magnetic force's direction and magnitude. The divergence operator, $\nabla \cdot$, measures how a vector field diverges from a point, and the curl operator, $\nabla \times$, assesses the field's rotation. Charge density is represented by $\rho$, influencing the electric field. The constants $\varepsilon_0$ and $\mu_0$ denote the vacuum's permittivity and permeability, fundamental to field propagation. The current density vector, $\mathbf{J}$, reflects charge flow per unit area.

To address the challenge of processing unstructured meshes from this dataset using CV-based models, we adopt a straightforward approach that involves padding and reshaping. Given the non-uniform number of mesh nodes across samples, we standardize the input by padding and reshaping it to fit the input dimension required by the CV-based models (256x256). This method, however, comes with a compromise: it neglects the spatial distribution of mesh nodes, which can vary from densely to sparsely populated regions. Consequently, reshaping into a 2D array fails to accurately represent the true neighborhood relationships among nodes. This simplification is a necessary trade-off when applying CV-based models to unstructured mesh configurations, reflecting the inherent limitations of adapting these methods to complex spatial data. We use the entire dataset which consists of 691 motor designs.

\subsection{Experimental setup}
    In the following we outline the methodology to evaluate the performance of six models across three distinct datasets with different mesh topographies: structured mesh for the Darcy dataset, graded mesh for the \emph{U-bend Flow} dataset, and unstructured mesh for the \emph{Electric Motor} dataset. We use 10\% of the total data for validation and another independent 10\% for testing the models after training.
    
    The study compares three CV-based and three graph-based models, maintaining constant hyperparameter across all experiments to maintain comparability. A uniform batch size of 4 is applied to every model, alongside a cosine learning rate scheduler with a warm-up phase, and an Adam optimizer. The optimizer uses a learning rate of 0.001 for graph-based models and 0.00001 for CV-based models. Specifically for the Dinov2 and DeiT models, the learning rate for the encoder is reduced by a factor of 100, and only the final two layers of the encoder are fine-tuned. This adjustment is made because large foundational models are prone to rapid unlearning if their learning rates are too large. For the sake of completeness, preliminary studies involved training the DeiT and DINOv2 models from scratch and fine-tuning all their layers. However, this approach's failure to yield improved outcomes leads us to exclude a detailed evaluation of these aspects from the discussion. The models are chosen to explore their performance in different mesh environments, with specific details provided in the model description subsections. 
    
    Model performance is assessed using two metrics: root mean square error (RMSE) and training time per epoch (TTE) for each combination of models and datasets. RMSE evaluates the models' performance based on ground-truth solutions from numerical PDE simulations, while the TTE measures their computational efficiency. These metrics allow for a balanced comparison of model capabilities across the datasets. Additionally, we conduct each experiment three times with varying random seeds, reporting both the mean $\mu$ and standard deviation $\sigma$ of the test RMSE. All experiments were conducted within an identical HPC cluster configuration, utilizing 16 cores of an AMD-EPYC 7743 CPU, 128GB of RAM, and a single NVIDIA-A100 GPGPU with 80GB of VRAM to provide a runtime comparison.

\section{Results and Discussion}\label{results}%

The results from all experiments are summarized in Table~\ref{tab:resultstable}, presenting the mean~$\mu$ and standard deviation~$\sigma$ of the RMSE for each model across three trials on various datasets. The analysis also includes the TTE and the number of trainable parameters for each model. Notably, for DeiT and DINOv2, the total number of parameters differs from the number of trainable parameters. The low standard deviation~$\sigma$ observed across all models on every dataset indicates successful experiments, indicating that the results are meaningful and not merely due to random variance.

\begin{table}[b!]
    \centering
    \caption{Experiment results comparing 6 models across 3 datasets, each trained thrice with different seeds. The table shows mean~$\mu$ and standard deviation~$\sigma$ for RMSE and TTE in seconds. Best performing models are highlighted in bold.}
    {\scriptsize%
    \begin{tabular}{|c|c|c|c|c|c|c|c||c|c|c|c|c|c|c|}
    \hline
    \multicolumn{2}{|c|}{base} & \multicolumn{6}{c||}{\textbf{CV}} & \multicolumn{6}{c|}{\textbf{Graph}} \\ 
    \hline
     \multicolumn{2}{|c|}{model}  & \multicolumn{2}{c|}{U-Net~\cite{ronneberger2015UNet}} & \multicolumn{2}{c|}{DINOv2~\cite{oquab2024dinov}} & \multicolumn{2}{c||}{DeiT~\cite{touvron2021training}} & \multicolumn{2}{c|}{BSMS~\cite{Cao2023Efficient}} & \multicolumn{2}{c|}{MGTN~\cite{ripken2023multiscale}} & \multicolumn{2}{c|}{MGN~\cite{pfaff2021learning}} \\ 
    \hline
    \multicolumn{2}{|c|}{n\_param}  & \multicolumn{2}{c|}{113M} & \multicolumn{2}{c|}{41M} & \multicolumn{2}{c||}{41M} & \multicolumn{2}{c|}{941K} & \multicolumn{2}{c|}{721K} & \multicolumn{2}{c|}{721K} \\ 
    \hline
    \multicolumn{2}{|c|}{metric}  & RMSE & TTE & RMSE & TTE & RMSE & TTE & RMSE & TTE & RMSE & TTE & RMSE & TTE \\
    \hline
    \hline
    \rowcolor{gray!10} Darcy~\cite{takamoto2022PDEBench} & $\mu$ & \textbf{6.61e-3} & 99  & 3.04e-2 & 24 & 3.22e-2 & 22 & 7.97e-3 & 298 & 1.88e-2 & 288 & 5.93e-2 & 246  \\
    \rowcolor{gray!10} (struc.) & $\sigma$ & \textbf{4.71e-4} & & 1.63e-3 & & 7.21e-4 & & 1.93e-4 & &6.45e-4 & & 2.96e-3 &  \\
    \hline
    \rowcolor{gray!30} U-bend~\cite{decke2023dataset} & $\mu$ & \textbf{7.89e-3} & 86 & 3.58e-2 & 14 & 6.13e-2 & 13 & 2.602e-2 & 1883 & 6.68e-2 & 1164 & 9.21e-2 & 71  \\
    \rowcolor{gray!30} (grad.) & $\sigma$  & \textbf{1.49e-4}& & 1.29e-3 & & 6.57e-4 & & 2.996e-3 & & 3.13e-3 & & 6.67e-3& \\
    \hline
    \rowcolor{gray!50} Motor~\cite{Botache2023Enhancing} & $\mu$ & 9.92e-2 & 94 & 2.74e-1 & 78 & 2.76e-1 & 73 & \textbf{8.17e-2} & 692 & 2.35e-1 & 529 & 2.45e-1 & 356 \\
    \rowcolor{gray!50} (unstr.) & $\sigma$  &7.02e-3 & & 3.02e-4 & & 2.52e-4 & & \textbf{3.34e-3} & & 1.13e-2 &  & 2.03e-2 & \\
    \hline
    \end{tabular}
    }%
    \label{tab:resultstable}

\end{table}

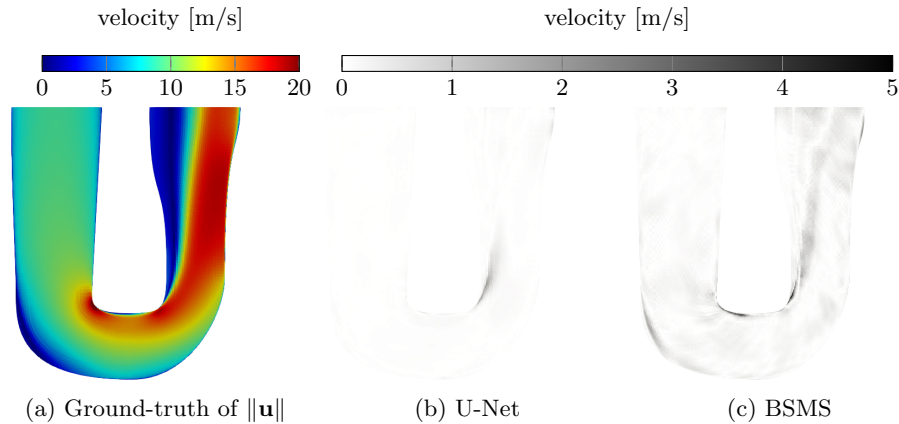
\begin{figure}[b!]
        \begin{tikzpicture}
            \begin{axis}[
                colorbar horizontal,
                colormap/jet,
                colorbar style={
                title={velocity [m/s]}, 
                width=0.28\textwidth, 
                height=6, 
                xtick={0,5,10,15,20},
                point meta min=0, 
                point meta max=20, 
                },
                paraviewStyle,
                ]
                \addplot3[
                    surf,
                    shader=interp,
                    domain=0:1,
                    y domain=0:1,
                    ] {20*x};
            \end{axis}
        \end{tikzpicture}
        \begin{tikzpicture}
            \begin{axis}[
                colorbar horizontal,
                colormap name=w_r,
                colorbar style={
                title={velocity [m/s]}, 
                width=0.6\textwidth, 
                height=6, 
                xtick={0,1,2,3,4,5},
                point meta min=0, 
                point meta max=5 
                },
                paraviewStyle,
                ]
                \addplot3[
                    surf,
                    shader=interp,
                    domain=0:1,
                    y domain=0:1,
                    ] {20*x};
            \end{axis}
        \end{tikzpicture}
        \centering
        \begin{subfigure}[b]{0.32\textwidth}
            \includegraphics[width=.8\textwidth]{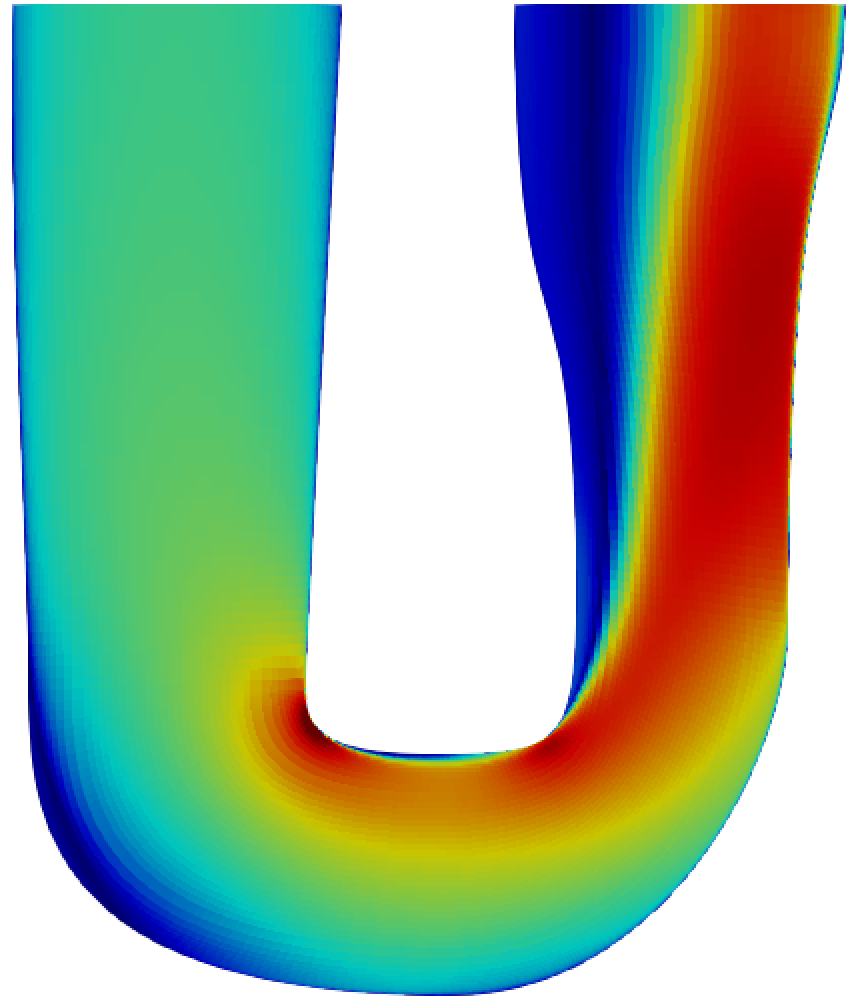}
            \caption{Ground-truth of $\lVert \mathbf{u} \rVert$}
            \label{fig:gt_u}
        \end{subfigure}
        \hfill
        \begin{subfigure}[b]{0.32\textwidth}
            \includegraphics[width=.8\textwidth]{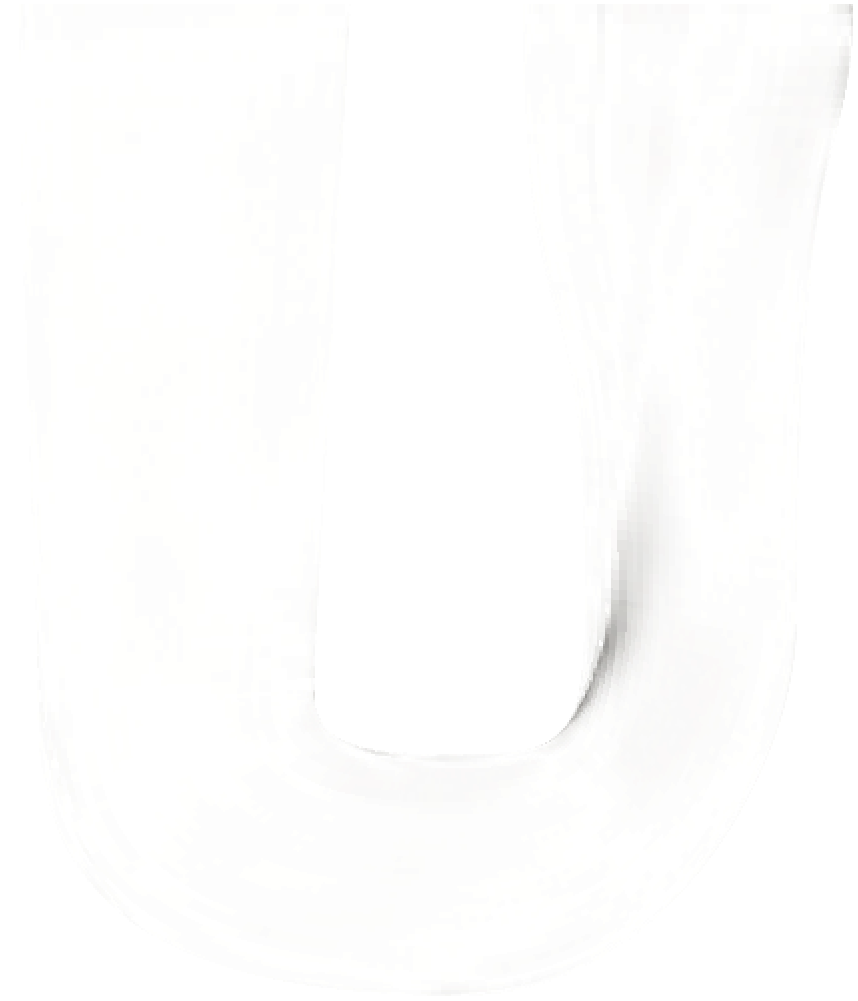}
            \caption{U-Net}
            \label{fig:unet_u}
        \end{subfigure}
        \hfill
        \begin{subfigure}[b]{0.32\textwidth}
            \includegraphics[width=.8\textwidth]{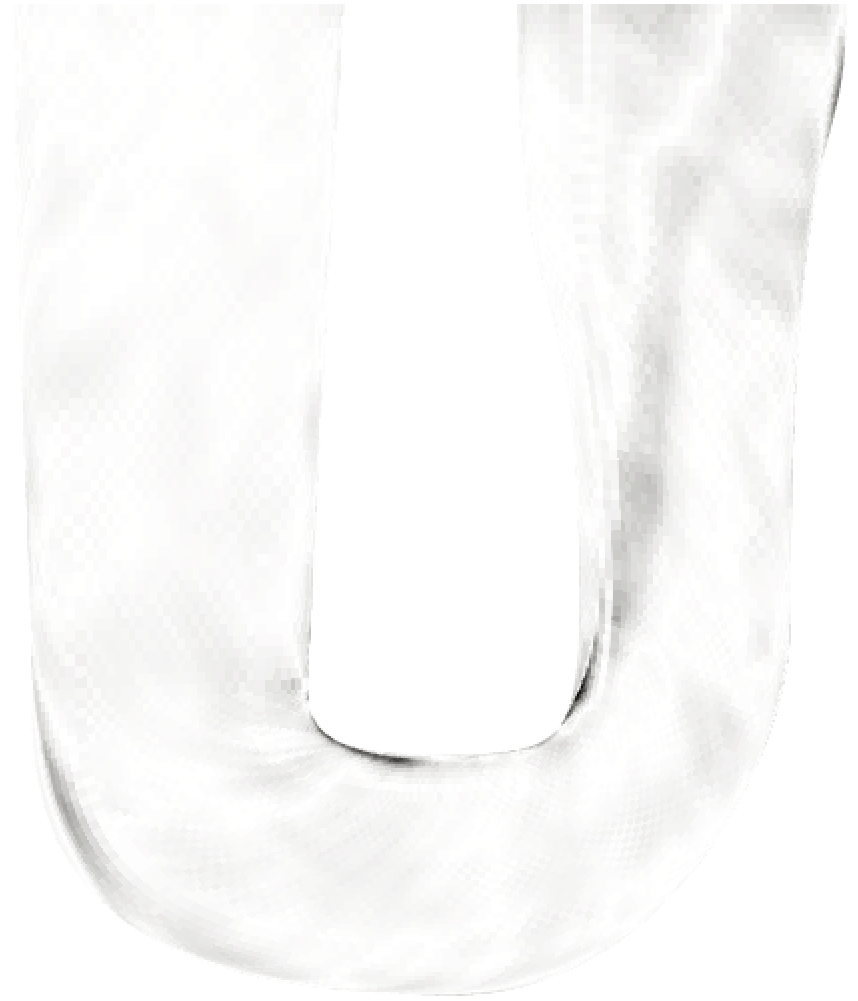}
            \caption{BSMS}
            \label{fig:bsms_u}
        \end{subfigure}
        \caption{Ground-truth of the \emph{velocity} $\lVert \mathbf{u} \rVert$ and the difference between ground-truth and the predictions of the best performing CV-based and graph-based model.}
        \label{fig:results_U_U-bend}
    \end{figure}

The top-performing model on each dataset is highlighted in bold. Specifically, within the domain of computer vision (CV) models attributing our contribution~\textbf{C1}, the U-Net model surpasses both DINOv2 and DeiT across all datasets, arranged by RMSE performance as U-Net $<$ DINOv2 $<$ DeiT. A similar consistent pattern is noted among the graph models, for contribution~\textbf{C2}, with RMSE performance ranked as BSMS $<$ MGTN $<$ MGN.
When comparing the highest-performing graph model (BSMS) against the top CV-based model (U-Net), to account our contribution~\textbf{C3}, it's observed that U-Net outperforms in the \emph{Darcy Flow} and \emph{U-bend Flow} datasets, while BSMS leads on the \emph{Electric Motor} dataset. The observed dominance of the U-Net model in processing the structured \emph{Darcy Flow} dataset, alongside a superior performance of the BSMS model on the unstructured mesh of the \emph{Electric Motor} dataset, aligns with our initial expectations. Nevertheless, the notable performance of U-Net over BSMS in evaluating the \emph{U-bend Flow} dataset, which includes graded meshes, is particularly engaging. This result implies that converting graded mesh cells into a 2D array, thus considering them as a regular mesh, notably helps in reducing the RMSE in model performance compared to treating them as graph data. Figure~\ref{fig:results_U_U-bend} displays the results of the \emph{velocity} vector~$\lVert \mathbf{u} \rVert$ for a sample from the \emph{U-bend Flow} dataset. Figure~\ref{fig:gt_u} shows the ground-truth, which is generated through time-intensive numerical simulations.

Figures~\ref{fig:unet_u} and~\ref{fig:bsms_u} present the absolut difference that illustrate the discrepancies between the models' predictions and the ground-truth for the U-net and the BSMS model, respectively. We use a grayscale colorbar where darker areas indicate larger errors in this specific cell location. In Figure~\ref{fig:unet_u}, it is observed that the U-net model exhibits a very low error across large areas of the solution space. The only exception is in the wake on the inner side of the channel, where flow separation would occur, and a vortex forms; here, the model slightly mispredicts. For the BSMS model in Figure~\ref{fig:bsms_u}, considerably larger errors are evident throughout the entire wake, as well as noticeable issues on the outer side at the inlet transition, highlighting the BSMS model's challenges in these areas.

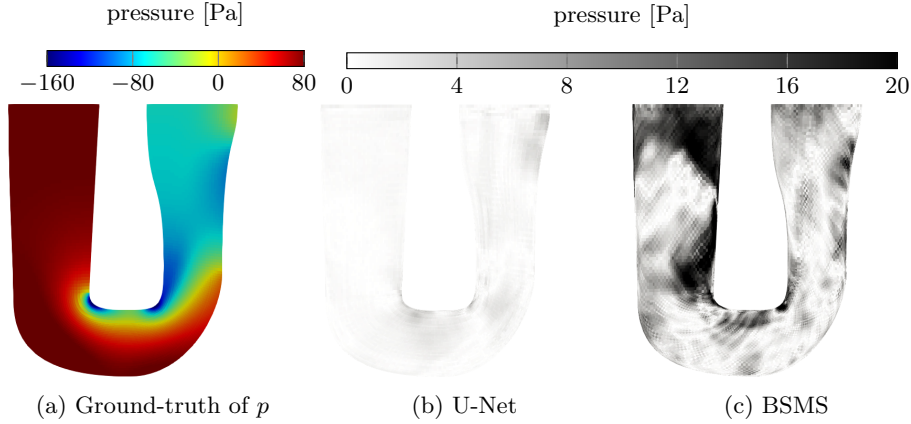
\begin{figure}[t!]
        \begin{tikzpicture}
            \begin{axis}[
                colorbar horizontal,
                colormap/jet,
                colorbar style={
                title={pressure [Pa]}, 
                width=0.28\textwidth, 
                height=6, 
                xtick={-160,-80,0,80},
                point meta min=-160, 
                point meta max=80 
                },
                paraviewStyle,
                ]
                \addplot3[
                    surf,
                    shader=interp,
                    domain=0:1,
                    y domain=0:1,
                    ] {20*x};
            \end{axis}
        \end{tikzpicture}
        \begin{tikzpicture}
            \begin{axis}[
                colorbar horizontal,
                colormap name=w_r,
                colorbar style={
                title={pressure [Pa]}, 
                width=0.6\textwidth, 
                height=6, 
                xtick={0,4,8,12,16,20},
                point meta min=0, 
                point meta max=20 
                },
                paraviewStyle,
                ]
                \addplot3[
                    surf,
                    shader=interp,
                    domain=0:1,
                    y domain=0:1,
                    ] {20*x};
            \end{axis}
        \end{tikzpicture}
        \centering
        \begin{subfigure}[b]{0.32\textwidth}
            \includegraphics[width=.8\textwidth]{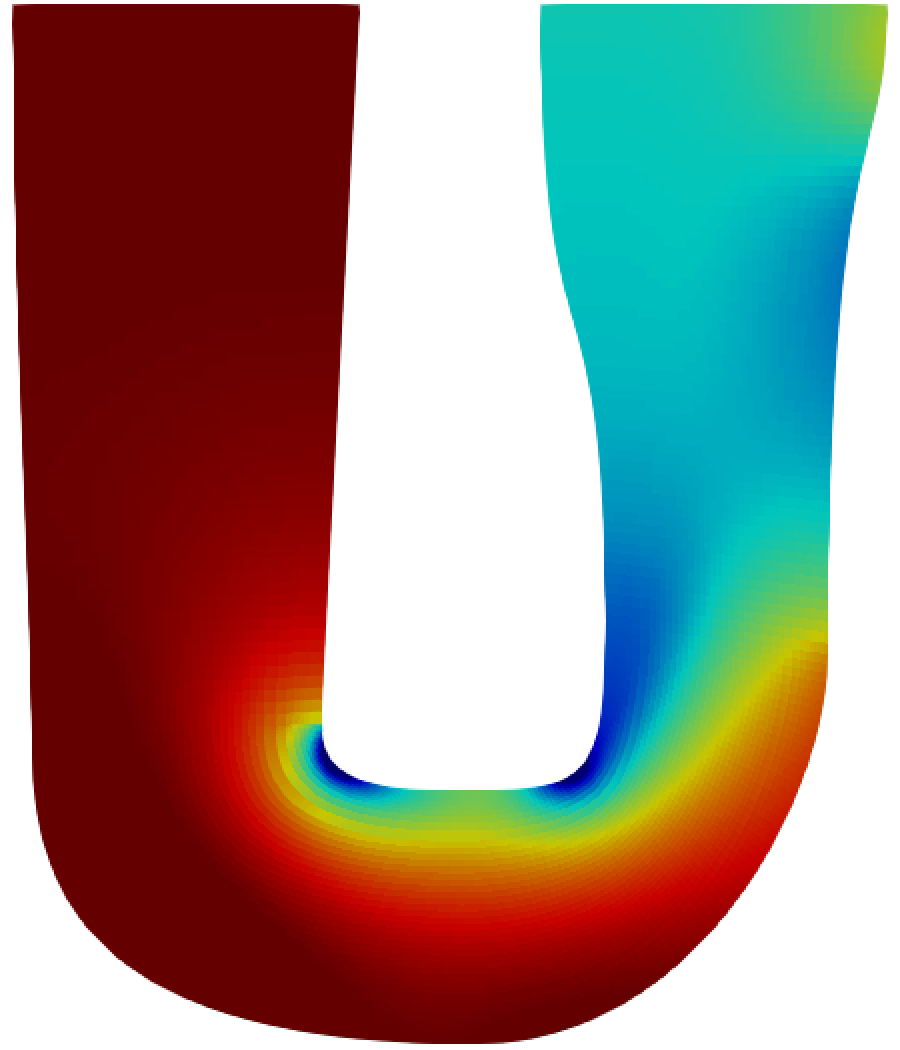}
            \caption{Ground-truth of $p$}
            \label{fig:gt_p}
        \end{subfigure}
        \hfill
        \begin{subfigure}[b]{0.32\textwidth}
            \includegraphics[width=.8\textwidth]{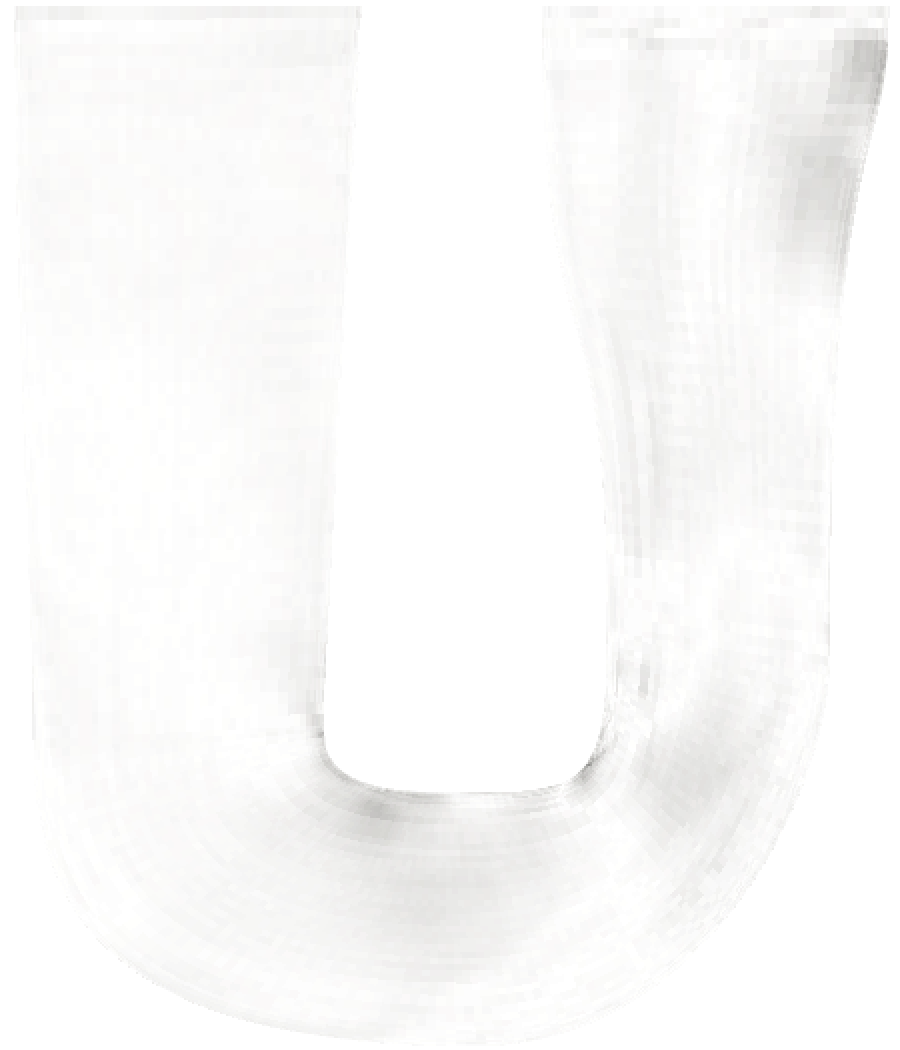}
            \caption{U-Net}
            \label{fig:unet_p}
        \end{subfigure}
        \hfill
        \begin{subfigure}[b]{0.32\textwidth}
            \includegraphics[width=.8\textwidth]{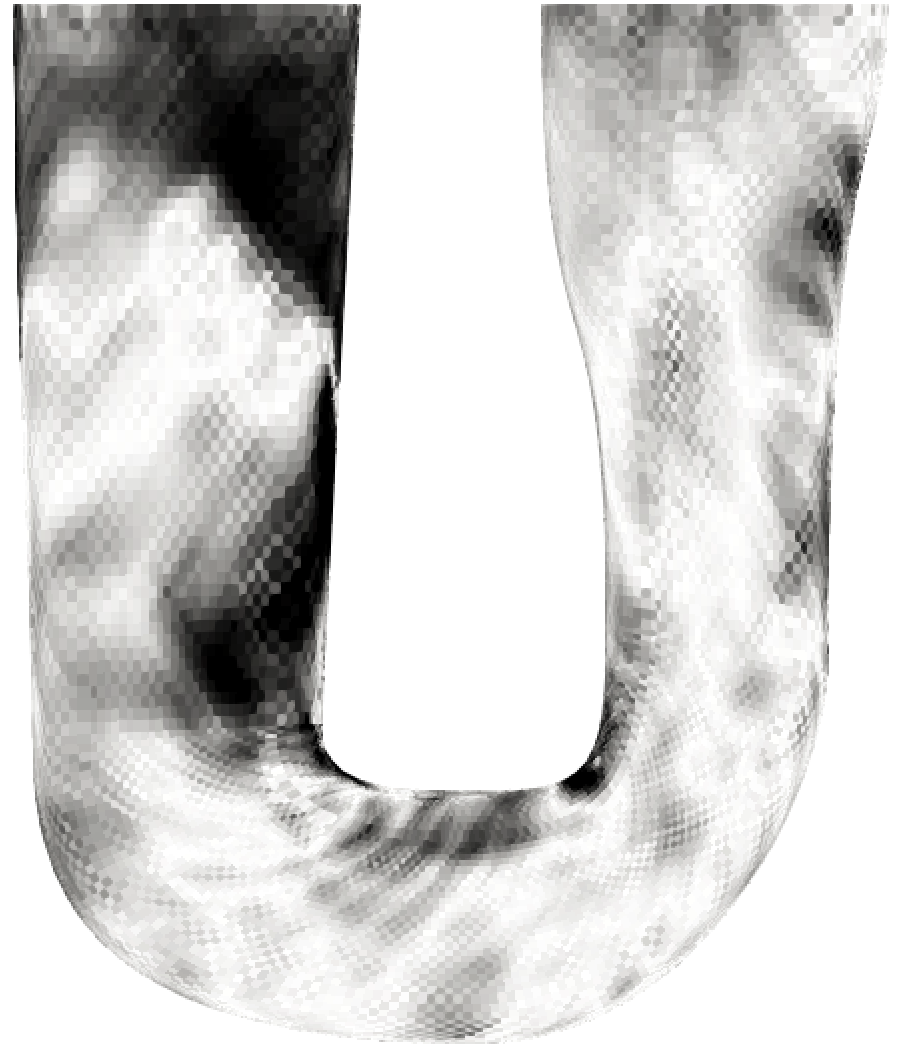}
            \caption{BSMS}
            \label{fig:bsms_p}
        \end{subfigure}
        \caption{Ground-truth of the \emph{pressure} $p$ and the difference between ground-truth and the predictions of the best performing CV-based and graph-based model.}
        \label{fig:results_p_U-bend}
    \end{figure}

Utilizing the \emph{U-bend Flow} dataset, we addressed two solution variables: in addition to the \emph{velocity} $\lVert \mathbf{u} \rVert$, we also solved for the \emph{pressure} $p$. In Figure~\ref{fig:results_p_U-bend} the ground-truth (cf. Figure~\ref{fig:gt_p}) and the two difference images in accordance to Figure~\ref{fig:results_U_U-bend} can be seen.

The observations made for Figure~\ref{fig:results_U_U-bend} are equally applicable to Figure~\ref{fig:results_p_U-bend}, with the findings being even more pronounced. The U-net model (cf. Figure~\ref{fig:unet_p}) provides more uniform estimation with lower error values. In addition to deviations at the transition point, discrepancies are also recorded on the outer side of the channel's wake. On the other hand, the deviations in the BSMS model (cf. Figure~\ref{fig:bsms_p}) appear vastly more irregular and thus less systematic or interpretable. Notably, the inlet is characterized by substantial deviations.

\begin{figure}[t!]
        \begin{tikzpicture}
            \begin{axis}[
                colorbar horizontal,
                colormap/jet,
                colorbar style={
                title={magnetic flux density [T]}, 
                width=0.28\textwidth, 
                height=6, 
                xtick={0,0.85,1.7,2.55},
                point meta min=0, 
                point meta max=2.55 
                },
                paraviewStyle,
                ]
                \addplot3[
                    surf,
                    shader=interp,
                    domain=0:1,
                    y domain=0:1,
                    ] {20*x};
            \end{axis}
        \end{tikzpicture}
        \begin{tikzpicture}
            \begin{axis}[
                colorbar horizontal,
                colormap name=w_r,
                colorbar style={
                title={magnetic flux density [T]}, 
                width=0.6\textwidth, 
                height=6, 
                xtick={0,0.1,0.2,0.3,0.4,0.5},
                point meta min=0, 
                point meta max=0.5 
                },
                paraviewStyle,
                ]
                \addplot3[
                    surf,
                    shader=interp,
                    domain=0:1,
                    y domain=0:1,
                    ] {20*x};
            \end{axis}
        \end{tikzpicture}
        \centering
        \begin{subfigure}[b]{0.32\textwidth}
            \includegraphics[width=.8\textwidth]{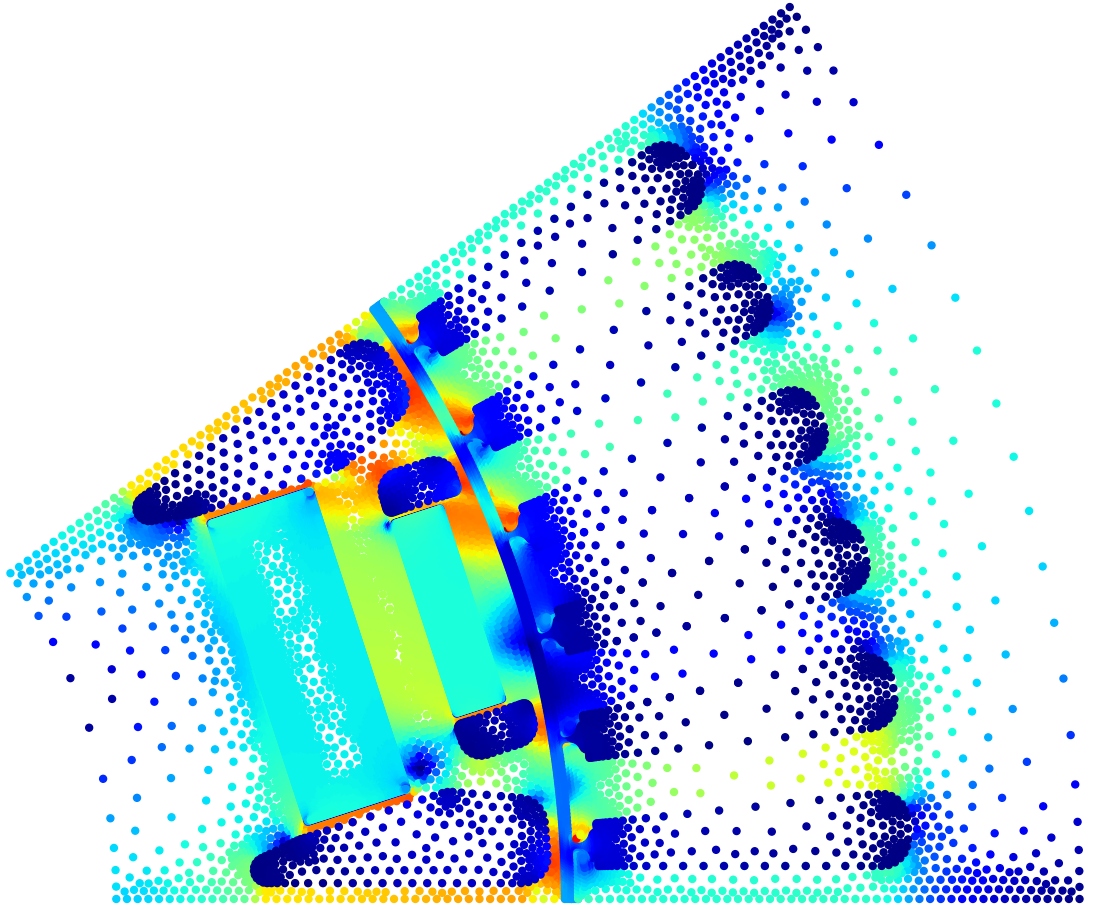}
            \caption{Ground-truth}
            \label{fig:gt_motor}
        \end{subfigure}
        \hfill
        \begin{subfigure}[b]{0.32\textwidth}
            \includegraphics[width=.8\textwidth]{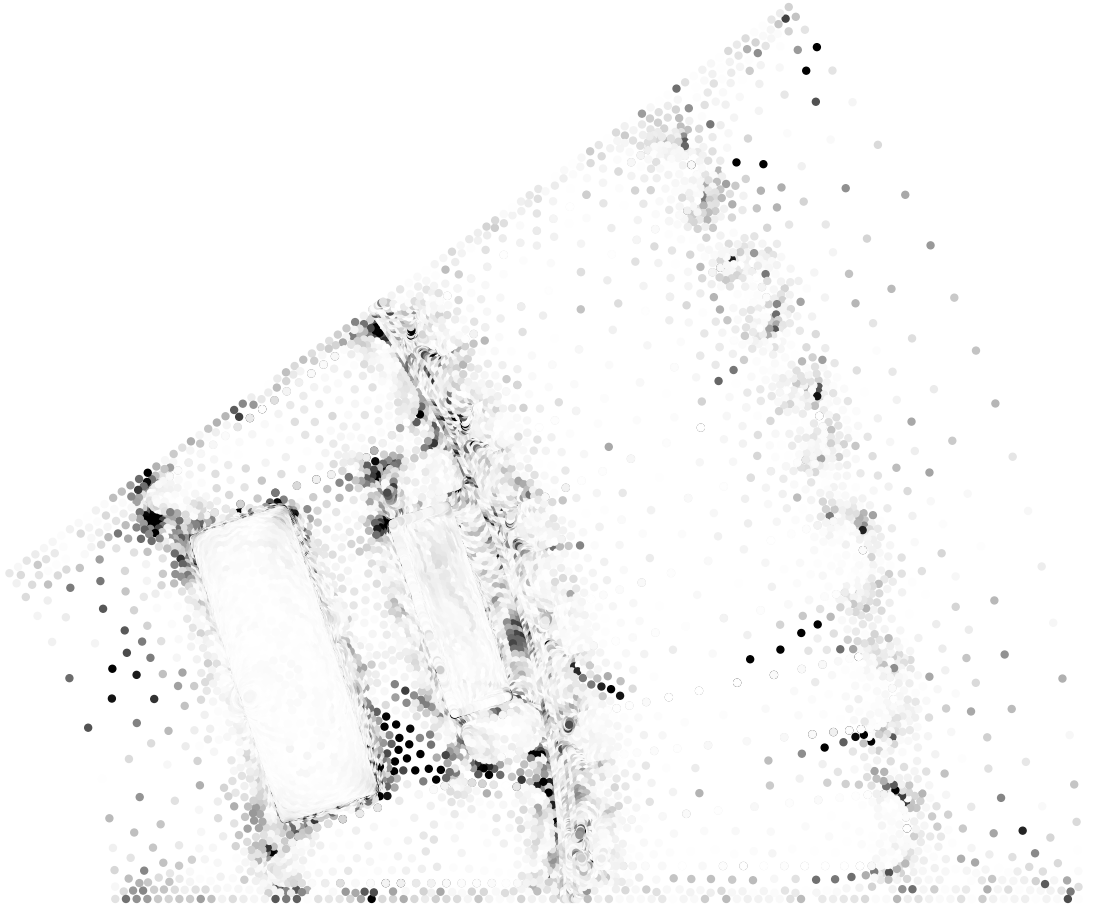}
            \caption{U-Net}
            \label{fig:unet_motor}
        \end{subfigure}
        \hfill
        \begin{subfigure}[b]{0.32\textwidth}
            \includegraphics[width=.8\textwidth]{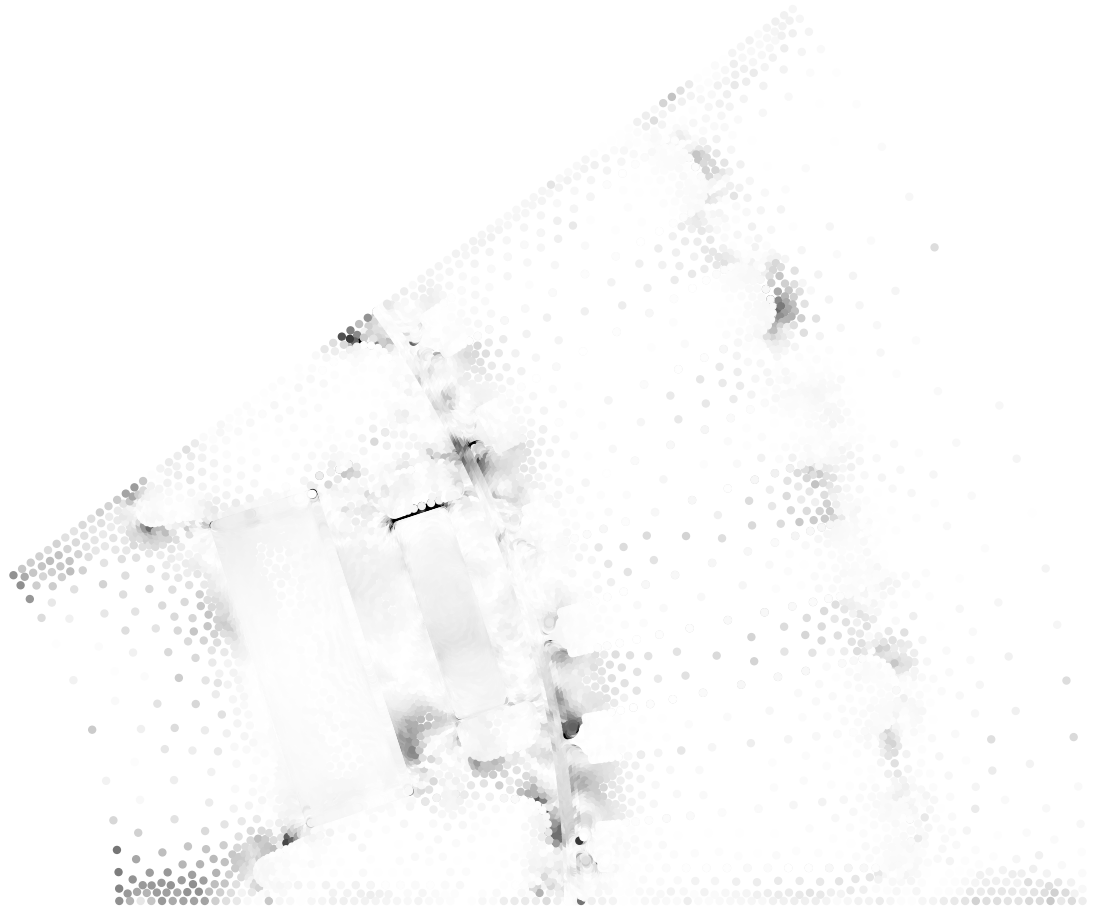}
            \caption{BSMS}
            \label{fig:bsms_motor}
        \end{subfigure}
        \caption{Ground-truth of \emph{magnetic flux density} $\mathbf{B}$ and difference between ground-truth and predictions of the best performing CV-based and graph-based model.}
        \label{fig:results_motor}
        \vspace{-10pt} 
    \end{figure}
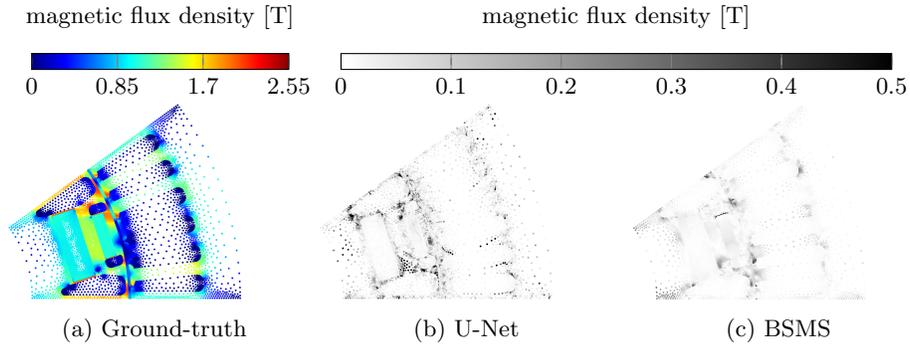

Contrary to expectations, the slight difference in performance between BSMS and U-Net on the \emph{Electric Motor} dataset is unexpected, revealing that our relatively straightforward strategy for managing unstructured mesh data with CV-based models can produce promising results. Figure~\ref{fig:results_motor} presents three images from a sample of the \emph{Electric Motor} dataset. Figure~\ref{fig:gt_motor} illustrates the ground-truth for the magnetic flux density $\mathbf{B}$. Figure~\ref{fig:unet_motor} displays the difference image, highlighting discrepancies between the ground-truth and the U-net model's prediction. Meanwhile, Figure~\ref{fig:bsms_motor} depicts the difference corresponding to the BSMS model's prediction. It is observable that there are no clear patterns for discrepancies within the solution space. Notably, the deviations of the U-Net are slightly larger than those of the BSMS model, a conclusion that was previously drawn from the analysis in Table~\ref{tab:resultstable} and is now further illustrated visually. To enhance the performance and applicability of predictions for electric motors, the computational mesh cells in the air gap between the rotor and stator are crucial for calculating motor torque for example. Employing a weighted loss function could underscore the importance of these cells in the mesh, thereby encouraging the model to focus on achieving more precise estimations in these critical areas. We omit a further evaluation of the \emph{Darcy Flow} dataset since all results in this dataset yielded the anticipated results.  Further experimental findings, along with the corresponding code, can be accessed via the link mentioned above.

When comparing the mean RMSE for contribution~\textbf{C3}, both the Dinov2 and the DeiT models surpass the MGN model in the \emph{Darcy Flow} and the \emph{U-bend Flow} datasets. However, they perform less effectively than the BSMS model. This under-performance may be attributed to the simplicity of the decoder models used, which might be inadequate for generating meaningful outputs from the encoded data. Another possible reason is that our specific application, such as solving mesh-based PDEs, presents an exceptionally challenging task for these pre-trained foundational models. Our approach involves fine-tuning the last two layers of these models only, as is common practice in CV. Yet, our use case deviates considerably from the original purposes and concepts behind these models. Developing foundation models tailored for solving mesh-based PDEs is an ambitious but potentially fruitful direction for future research. 

This assertion is further supported by examining the TTE, where the Dinov2 and DeiT models demonstrate the shortest training durations due to their fine-tuning on only the last two layers. Specifically, for the \emph{U-bend Flow} dataset, both models surpass two state-of-the-art graph models (MGN, MGTN) in their current simplistic forms. The presence of a large, pre-trained foundation model, coupled with an enhanced decoder network, holds meaningful potential for future work. In response to our contribution~\textbf{C4}, a short TTE is especially crucial for tasks such as DADO, where the primary goal is to quickly achieve a correct ranking of samples i.e. design candidates rather than obtaining the most precise results in a notably higher amount of time. Additionally, in domains involving mesh-based PDEs where a balance between TTE and performance must be struck, our comparison provides a valuable guide for potential model selection. 

\section{Conclusion}\label{conclusion}%
This article explores the use of deep learning as a more sustainable alternative to traditional HPC methods for solving PDEs. Our research investigates advanced CV-based and graph-based models, applied to three mesh topographies: structured, graded, and unstructured. We compare three CV-based models and three graph-based models, focusing on performance and computational efficiency across three different datasets. Here, we address the contributions outlined in Section~\ref{introduction}:
\begin{itemize}
    \item[] \textbf{C1:} We have validated the effectiveness of CV-based models such as the U-Net as well as the DINOv2 and DeiT foundation models, across three mesh configurations. These models are traditionally favored for structured meshes. A simple decoder network was developed for the foundation models, leveraging learned representations to generate predictions. To process data from unstructured meshes, we employed a straightforward strategy that converts the data into a 2D array, enabling its compatibility with CV-based models for processing.
    \item[] \textbf{C2:} We examined the performance of three graph-based models, the go-to approach for unstructured meshes. Therefore, existing implementations of the models have been adapted and applied to our experimental setup.
    \item[] \textbf{C3:} A key novelty of our work lies in the exploration of graded meshes, a mesh topography that had not been directly compared before. Our analysis revealed that CV-based models not only adapt well across all models but also outperform graph-based models in terms of both prediction performance and training speed considering structured and graded meshes. Noteworthy is the unexpected effectiveness of CV-based models in dealing with unstructured meshes, a domain traditionally dominated by graph-based models. 
    \item[] \textbf{C4:} The integration of foundation models like DINOv2 and DeiT represents a paradigm shift in DADO, highlighting the use of surrogate models for instance ranking. DADO prioritizes rapid assessment, which is then complemented by an expert reevaluation through numerical PDE simulation to get the ground-truth of selected instances.
    
    In contrast, SAM relies entirely on the models' outputs, bypassing expert corrections. This prioritizes the use of high-performing models such as U-Net and BSMS for SAM, underscoring their importance in delivering reliable predictions for autonomous operations. 
\end{itemize}
Given the surprising success of CV-based models, driven by a simple decoder and straightforward data handling, future efforts will focus on decoder optimization. Introducing skip connections between the encoder and decoder could enhance results by merging U-Net key features with advanced vision transformers. Moreover, adopting a refined strategy for processing unstructured mesh data could further improve outcomes. The development of foundation models, specifically trained on PDE data, holds the potential to markedly boost performance while substantially shortening training times.

Our study highlights the pivotal role of careful model selection in solving mesh-based PDEs, noting how methodologies like DADO and SAM may favor different models due to their requirements. This distinction underlines the tailored application of surrogate models to fulfill specific objectives. Our research advances computational methodologies and signals the development of more flexible, efficient, and environmentally friendly approaches for tackling complex PDE problems. The promising future of deep learning in this area points to computational efficiency and sustainability as key scientific inquiry alongside the development of organic computing methodologies.
\vspace{-0.425cm}
\bibliographystyle{ieeetr}
\bibliography{bibfile}
\end{document}